\documentclass{article}

\usepackage{microtype}
\usepackage{graphicx}
\usepackage{subcaption}
\usepackage{booktabs} 
\usepackage{enumitem} 
\usepackage{algorithm} 
\usepackage{algorithmicx} 
\usepackage{algpseudocode}
\usepackage{xcolor}

\PassOptionsToPackage{hyphens}{url}
\usepackage{hyperref}

\usepackage[preprint]{icml2026}

\usepackage{amsmath}
\usepackage{amssymb}
\usepackage{mathtools}
\usepackage{amsthm}
\usepackage[capitalize,noabbrev]{cleveref}

\theoremstyle{plain}

\theoremstyle{definition}

\theoremstyle{remark}

\usepackage[textsize=tiny]{todonotes}

\begin{document}

\twocolumn[
  \icmltitle{MIRL: Mutual Information-Guided Reinforcement Learning for Vision-Language Models}



  \icmlsetsymbol{equal}{*}

    \begin{icmlauthorlist}
      \icmlauthor{Yin Zhang}{equal,tju}
      \icmlauthor{Jiaxuan Zhao}{equal,iie}
      \icmlauthor{Zonghan Wu}{ecnu}
      \icmlauthor{Zengxiang Li}{enn}
      \icmlauthor{Junfeng Fang}{nus}
      \icmlauthor{Kun Wang}{ntu}
      \icmlauthor{Qingsong Wen}{squirrel}
      \icmlauthor{Yilei Shao}{ecnu}
    \end{icmlauthorlist}

    \icmlaffiliation{tju}{School of Mathematics, Tianjin University, Tianjin, China}
    \icmlaffiliation{iie}{Institute of Information Engineering, Chinese Academy of Sciences, Beijing, China}
    \icmlaffiliation{ecnu}{Shanghai Advanced Institute of Finance (SAIFS), East China Normal University, Shanghai, China}
    \icmlaffiliation{enn}{ENN Group, Digital Technology Research Institute, China}
    \icmlaffiliation{ntu}{Nanyang Technological University, Singapore}
    \icmlaffiliation{squirrel}{Squirrel AI Learning, Shanghai, China}
    \icmlaffiliation{nus}{National University of Singapore, Singapore}

    \icmlcorrespondingauthor{}{}

  \icmlkeywords{Vision-Language Models, Reinforcement Learning, Mutual Information, Efficient Sampling}

  \vskip 0.3in
]

\fancyhead[C]{MIRL: MI-Guided Reinforcement Learning}



\printAffiliationsAndNotice{\icmlEqualContribution}  

\begin{abstract}
\noindent

Vision-Language Models (VLMs) frequently suffer from visual perception errors and hallucinations that compromise answer accuracy in complex reasoning tasks.
Reinforcement Learning with Verifiable Rewards (RLVR) offers a promising solution by optimizing policies using answer correctness signals.
Despite their effectiveness, prevailing RLVR methods face two critical limitations.
First, much of the sampling budget is wasted on trajectories doomed to fail due to early visual description errors.
Second, sparse rewards cannot distinguish whether failures stem from visual perception or reasoning stages.
We introduce \textsc{MIRL}, a decoupled framework that addresses both limitations by leveraging mutual information (MI) between generated descriptions and visual inputs as a cheap pre-screening signal.
This enables intelligent budget allocation toward high-potential trajectories via forking, while decoupled training provides independent MI-based rewards for visual perception optimization, resolving reward blindness.
Experiments on six vision-language reasoning benchmarks demonstrate that \textsc{MIRL} achieves 70.22\% average accuracy and successfully surpasses the performance of sampling 16 complete trajectories using only 10 pre-samples with top-6 selection (25\% fewer complete trajectories).
Our code is available at: \url{https://anonymous.4open.science/r/mirl-main/}.

\end{abstract}

\section{Introduction}
\label{sec:introduction}

Vision-Language Models (VLMs) have demonstrated remarkable multimodal understanding and reasoning capabilities through large-scale pretraining, achieving significant progress on vision-intensive tasks such as chart analysis and scientific figure comprehension\citep{DBLP:conf/nips/Dai0LTZW0FH23, DBLP:conf/nips/LiuLWL23a, DBLP:journals/corr/abs-2308-12966}. However, they frequently suffer from visual perception errors, generating text inconsistent with visual facts, which compromises the accuracy of answers in complex visual reasoning tasks.

Reinforcement Learning with Verifiable Rewards (RLVR) offers a promising end-to-end solution for mitigating VLM hallucinations without external annotations, directly optimizing policy networks using verifiable signals such as answer correctness\citep{DBLP:journals/corr/abs-2402-03300, DBLP:journals/corr/abs-2503-14476}. Yet efficiently leveraging reward signals for fine-grained visual perception optimization remains an open challenge. Existing RLVR approaches fall into two categories based on reward design: (1) sparse reward methods use final answer correctness as the sole supervision signal, optimizing policies via group-relative advantage estimation\citep{DBLP:journals/corr/abs-2402-03300, DBLP:journals/corr/abs-2503-14476, DBLP:journals/corr/abs-2503-06749}, and (2) process reward methods introduce external Process Reward Models (PRMs) to evaluate intermediate reasoning steps, providing more granular supervision \citep{DBLP:conf/iclr/LightmanKBEBLLS24, DBLP:conf/acl/WangLSXDLCWS24}.

\begin{figure*}[ht]
  \vskip 0.2in
  \begin{center}
    \centerline{\includegraphics[width=\textwidth]{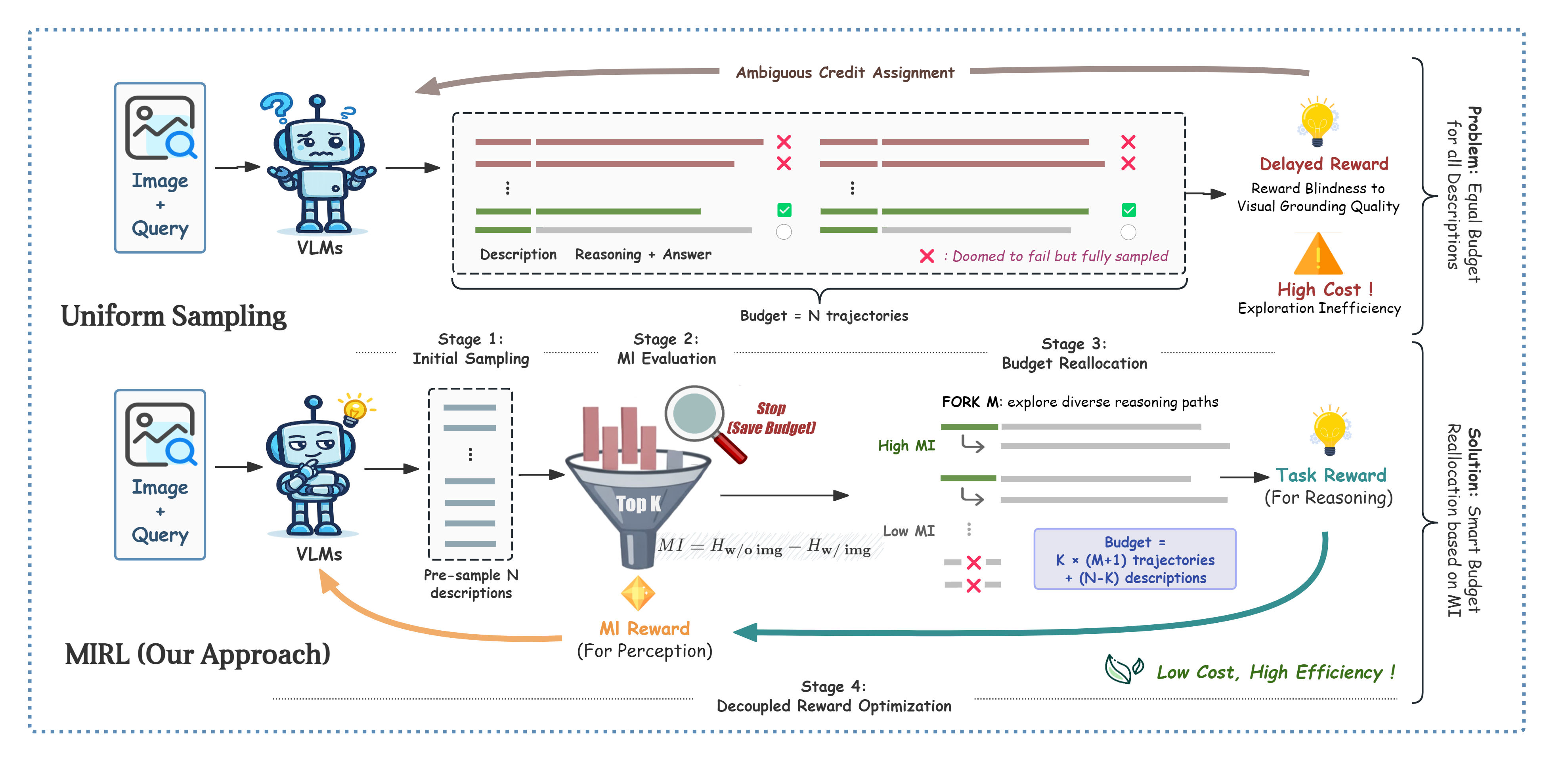}}
    \caption{\textbf{Top:} Existing methods allocate uniform sampling budget across all trajectories regardless of description quality, wasting resources on trajectories doomed to fail due to early visual perception errors. \textbf{Bottom:} MIRL first performs cheap pre-sampling, then uses Mutual Information (MI) to identify high-quality descriptions and conducts additional exploration via ``forking''; low-MI descriptions are retained only as negative samples.}
    \label{fig:comparison}
  \end{center}
  \vskip -0.2in
\end{figure*}

Despite their theoretical soundness, these methods face two critical limitations.

\begin{itemize}[leftmargin=2em, label=$\bullet$]
    \item \textbf{Inefficient Exploration from Uniform Sampling.} 
    Generating complete reasoning sequences incurs massive computational cost, making sampling prohibitively expensive. Yet much of this budget is wasted on trajectories that are ``doomed-to-fail''. If the visual description $v$ contains critical information loss or recognition errors, no amount of sophisticated reasoning can later recover the correct answer.
    
    \item \textbf{Reward Blindness to Visual Grounding Quality.} As illustrated in Figure~\ref{fig:comparison} (top), when a complete trajectory fails, the task reward provides only a binary signal at the sequence level. Policy updates cannot distinguish whether the failure stems from visual perception, reasoning logic, or answer formulation. This credit assignment ambiguity severely hinders targeted optimization of individual generation stages.
\end{itemize}

To simultaneously address sampling efficiency and reward blindness, we draw inspiration from M3ID~\citep{DBLP:conf/cvpr/FaveroZTCPASS24} and CMI-VLD~\citep{DBLP:journals/corr/abs-2505-19678}, which successfully suppresses VLM hallucinations by maximizing MI between generated tokens and visual inputs during inference. We migrate mutual information (MI) from inference to training, proposing a strategy that kills two birds with one stone.

Rather than equally allocating sampling budget across all trajectories, we intelligently concentrate resources on high-potential ones. \textbf{For the first limitation}, we introduce MI as a cheap ``pre-screening'' signal to evaluate visual description quality at low cost. Leveraging MI's ability to measure how strongly generated content depends on image information, we rapidly evaluate and filter numerous low-cost description candidates from the policy network for budget reallocation. The top-$K$ high-MI descriptions undergo additional exploration via ``forking'', while low-MI descriptions are excluded from subsequent reasoning sampling.

Then, \textbf{for the second limitation}, we propose a decoupled training strategy. High-MI trajectories undergo complete sampling and receive policy updates based on task rewards. All pre-sampled descriptions receive immediate MI-based reward signals for updates. As shown in Figure~\ref{fig:comparison} (bottom), this decoupling mechanism not only improves sampling efficiency but also provides independent, low-latency reward signals for visual description generation, effectively resolving the credit assignment problem. Detailed methodology and theoretical analysis are provided in \cref{sec:method}. We term this approach \textbf{MIRL}, \textbf{M}utual \textbf{I}nformation-guided \textbf{R}einforcement \textbf{L}earning, an efficient decoupled VLM training method guided by mutual information.

Our validation reveals several practically valuable properties. Regarding generalization, MI signals exhibit stable indicative power across diverse task domains including mathematical reasoning and general visual QA, demonstrating universal value as a proxy for visual perception quality. Ablation studies show that using MI rewards alone has clear limitations, performing significantly worse than synergistic combination with task correctness rewards. This validates the necessity of decoupling where the perception stage requires independent signals while reasoning still depends on supervision from final answers. 
Regarding engineering implementation, MIRL adopts a modular design by introducing forking sampling logic in the Rollout stage and embedding decoupled reward functions during loss computation. It can be compatibly integrated with mainstream RL algorithms such as GRPO~\citep{DBLP:journals/corr/abs-2402-03300} and DAPO~\cite{DBLP:journals/corr/abs-2503-14476} without restructuring the core optimization framework.

We conduct extensive experiments on multiple VLMs reasoning benchmarks including \textbf{MathVista}~\citep{DBLP:conf/iclr/LuBX0LH0CG024}, \textbf{MathVerse}~\citep{DBLP:conf/eccv/ZhangJZLGQZLCQGL24}, and \textbf{We-Math} for mathematical reasoning, \textbf{M3CoT} for multi-modal chain-of-thought, \textbf{MMStar}~\citep{DBLP:conf/nips/ChenLDZZCDWQLZ24} and \textbf{RealWorldQA}~\citep{xai2024realworldqa} for general vision-language understanding. With 10 pre-sampled descriptions, selecting top-6 by MI for complete sampling with 12 trajectories each, MIRL achieves convergence speed on validation accuracy over direct sampling of 16 complete trajectories, while surpassing its performance with computational cost reduction by approximately 25\%.
This work reveals the significant value of mutual information as an intermediate evaluation signal in VLM reinforcement learning, opening new research directions for efficient large-scale VLM training.

\section{Preliminary}
\label{sec:preliminary}

\subsection{Vision-Language RL with Verifiable Rewards}
\label{subsec:prelim_rlvr}
Consider a vision-language reasoning task where each instance comprises an image $I$, a textual query $q$, and a verifiable ground-truth answer $a^*$.
A vision-language model (VLM) with parameter $\theta$ defines an autoregressive policy $\pi_\theta$ that generates a response $y = (y_1, \ldots, y_T)$ token by token:
\begin{equation}
\pi_\theta(y \mid I, q) = \prod_{t=1}^{T} \pi_\theta(y_t \mid I, q, y_{<t}).
\label{eq:autoregressive}
\end{equation}
Reinforcement Learning with Verifiable Rewards (RLVR) optimizes this policy by maximizing expected task correctness.
Upon generating a complete response, the model receives a sparse reward $r_{\text{task}}(y, a^*) = \mathbb{I}[\texttt{extract}(y) = a^*]$.
Policy updates proceed via group-relative advantage estimation\citep{DBLP:journals/corr/abs-2402-03300}: for each query, $G$ responses $\{y^{(i)}\}_{i=1}^G$ are sampled and advantages computed as

\begin{equation}
\hat{A}^{(i)} = \frac{r^{(i)} - \mu_G}{\sigma_G},
\label{eq:grpo_advantage}
\end{equation}
where $\mu_G = \frac{1}{G}\sum_{j} r^{(j)}$ and $\sigma_G = \sqrt{\frac{1}{G}\sum_{j} (r^{(j)} - \mu_G)^2}$.

\subsection{Visual Perception as the Bottleneck}
\label{subsec:prelim_bottleneck}

In vision-language reasoning, generated trajectories typically follow an information flow from visual perception to abstract reasoning: the model first extracts and describes image content, then reasons over this description to derive an answer.
Let $y = (\mathcal{D}, \mathcal{R}, a)$ denote a trajectory, where $\mathcal{D}$ represents the early description segment grounded in visual input, $\mathcal{R}$ the subsequent reasoning, and $a$ the final answer.

This sequential structure implies a bottleneck: if $\mathcal{D}$ fails to capture task-relevant visual information, subsequent reasoning operates on an impoverished representation and is unlikely to recover the correct answer.
By analogy to the data processing inequality, information about $I$ available at the answer stage is bounded by what is preserved in the description:
\begin{equation}
I(a; I \mid q) \leq I(\mathcal{D}; I \mid q).
\label{eq:dpi}
\end{equation}

This suggests that description quality, quantified through mutual information, could predict downstream success without requiring full trajectory generation.

\subsection{Mutual Information for Assessing Visual Dependence}
\label{subsec:prelim_mi}

To quantify how much a generated description depends on visual input, we measure MI between the description tokens and the image.
For each token $y_t$ in the generation process, the MI with respect to image $I$
 is defined as
\begin{equation}
\text{MI}(y_t; I \mid q, y_{<t}) = H(y_t \mid q, y_{<t}) - H(y_t \mid
 I, q, y_{<t}),
\label
{eq:mi}
\end{equation}
where $H(\cdot)$
 denotes conditional entropy. High MI indicates the model is generating image-dependent content; low MI suggests reliance on textual priors, signaling potential perception failures.

For a description segment $\mathcal{D} = (y_1, \ldots, y_k)$, we aggregate token-level MI via averaging:
\begin{equation}
\text{MI}(\mathcal{D}; I \mid q) = \frac{1}{k} \sum_{t=1}^{k} \text{MI}(y_t; I \mid q, y_{<t}).
\label{eq:segment_mi}
\end{equation}

Computing \cref{eq:segment_mi} requires estimating two conditional entropy terms via forward passes through the model, substantially more efficient than generating full reasoning trajectories. 
This enables MI as an early quality indicator for budget allocation, as we demonstrate in \cref{subsec:method_mi_indicator,subsec:method_allocation}.

\section{Method}
\label{sec:method}

We present \textsc{MIRL} (\textbf{M}utual \textbf{I}nformation-guided \textbf{R}einforcement \textbf{L}earning), which leverages MI as a cheap pre-screening signal to improve sampling efficiency and credit assignment in vision-language RL.
The core contribution is MI-guided sampling budget allocation; the MI-based reward serves as a complementary signal.

\subsection{MI as Early Indicator of Trajectory Success}
\label{subsec:method_mi_indicator}

We first validate that MI predicts trajectory success by examining the relationship between description-level MI and task accuracy.
For each benchmark, we sample multiple trajectories per question, compute MI scores via \cref{eq:segment_mi}, partition trajectories into MI-based bins, and measure accuracy within each bin.

As shown in \cref{fig:mi_correlation}, we observe consistent positive correlation across diverse benchmarks spanning mathematical reasoning (MathVista, MathVerse, We-Math), scientific problem solving (M3CoT), and general visual understanding (RealWorldQA, MMStar).
Pearson correlation coefficients range from $r=0.71$ to $r=0.94$, with all benchmarks achieving statistical significance at $p<0.05$.
On MathVista, accuracy increases from approximately 32\% in the lowest MI bin to 80\% in the highest, corresponding to a 146.9\% relative improvement.

To ensure MI genuinely measures visual grounding rather than language-only shortcuts, we filter out samples where the model achieves correct answers without visual input, retaining only vision-dependent questions for this analysis.
This confirms that MI computed from $\mathcal{D}$ alone reliably predicts downstream task success, enabling budget allocation decisions immediately after description generation without requiring full trajectory completion.


\subsection{MI-Guided Sampling Budget Allocation}
\label{subsec:method_allocation}

Rather than uniformly sampling $G$ complete trajectories, \textsc{MIRL} 
employs a \textbf{sample-evaluate-fork} strategy that intelligently concentrates budget on high-potential descriptions.

\textit{Phase 1: Description Pre-Sampling.}
Generate $N$ response candidates from the policy $\pi_\theta(\cdot \mid I, q)$.
Each candidate contains a description segment that will be evaluated for 
quality before allocating reasoning exploration budget.

\textit{Phase 2: MI-Based Selection.}
For each candidate, compute the MI score between its description $\mathcal{D}^{(i)}$ and the visual input:
\begin{equation}
s^{(i)} = \text{MI}(\mathcal{D}^{(i)}; I \mid
 q)
\end{equation}

Select the top-$K$ candidates:
\begin{equation}
\mathcal{S}_K = \textsc{TopK}\bigl(\{s^{(i)}\}_{i=1}^{N}, K\bigr
)
\label
{eq:selection}
\end{equation}

\textit{Phase 3: Budget Reallocation via Forking.}
For each selected description $\mathcal{D}^{(i)}$ where $i \in \mathcal{S}_K$, we generate $(M+1)$ complete trajectories: the original $y^{(i)} = (\mathcal{D}^{(i)}, \mathcal{R}^{(i)}, a^{(i)})$ plus $M$ forked branches $\tilde{y}^{(i,j)} = (\mathcal{D}^{(i)}, \tilde{\mathcal{R}}^{(i,j)}, \tilde{a}^{(i,j)})$ for $j = 1, \ldots, M$, where forked reasoning is sampled as $\tilde{\mathcal{R}}^{(i,j)}, \tilde{a}^{(i,j)} \sim \pi_\theta(\cdot \mid I, q, \mathcal{D}^{(i)})$. This creates a tree structure with high-MI descriptions as roots and multiple reasoning branches. 

For unselected descriptions $i \notin \mathcal{S}_K$, only $\mathcal{D}^{(i)}$ is retained for perception training. The final training set contains $K(M+1)$ complete trajectories and $(N-K)$ descriptions.

\subsection{Decomposed Reward Assignment}
\label{subsec:method_training}

Beyond budget allocation via MI-selection and forking,  \textsc{MIRL} addresses reward blindness by providing separate reward signals for visual perception and reasoning.

\textbf{MI-Based Reward.}
All $N$ pre-sampled descriptions receive an immediate reward $r_{\text{MI}}^{(i)} = \text{MI}(\mathcal{D}^{(i)}; I \mid q)$, regardless of selection status.
This provides dense, low-latency feedback for description generation.
Empirically, this signal primarily stabilizes training and accelerates format alignment rather than directly improving accuracy in isolation.

\textbf{Task-Based Reward.}
$K(M+1)$ complete trajectories from selected descriptions receive standard task rewards $r_{\text{task}}^{(i,j)} = \mathbb{I}[\texttt{extract}(y^{(i,j)}) = a^*]$ for $i \in \mathcal{S}_K$ and $j = 0, \ldots, M$
, which supervise reasoning and answer formulation.

\textbf{Decoupled Updates.}
Advantages are computed separately: MI-based advantages $\hat{A}_{\text{MI}}^{(i)}$ for description tokens and task-based advantages $\hat{A}_{\text{task}}^{(i,j)}$ for full trajectories, both following the group-relative formulation in \cref{eq:grpo_advantage}.
This ensures visual perception optimization receives independent feedback unconfounded by reasoning quality.

The complete procedure is summarized in \cref{alg:mirl}.
\textsc{MIRL} integrates with existing RLVR algorithms (GRPO, DAPO) by modifying only the sampling strategy and adding MI rewards, without changes to underlying policy gradient computation.

\begin{algorithm}[t]
\caption{\textsc{MIRL}: MI-Guided Reinforcement Learning}
\label{alg:mirl}
\begin{algorithmic}[1]
\Require Policy $\pi_\theta$, dataset $\mathcal{D}_{\text{train}}$
\Require $N$, $K$, $M$: sampling hyperparameters

\For{each training iteration}
    \For{each prompt $(I, q, a^*)$ from $\mathcal{D}_{\text{train}}$}
        \State \textcolor{gray}{// Phase 1: Initial sampling}
        \State Generate $N$ candidates: $\{\mathcal{D}^{(i)}\}_{i=1}^{N}$
        
        \State \textcolor{gray}{// Phase 2: MI-guided selection}
        \State Compute $s^{(i)} \gets \text{MI}(\mathcal{D}^{(i)}; I \mid q)$ for all $i$
        \State Select top-$K$: $\mathcal{S}_K \gets \textsc{TopK}(\{s^{(i)}\}, K)$
        
        \State \textcolor{gray}{// Phase 3: Differentiated trajectory exploration}
        \For{$i \in \mathcal{S}_K$}
            \State Complete original trajectory: $y^{(i)} = (\mathcal{D}^{(i)}, \mathcal{R}^{(i)}, a^{(i)})$
            \State Fork $M$ new trajectories from $\mathcal{D}^{(i)}$
        \EndFor
        
        \State \textcolor{gray}{// Phase 4: Decomposed reward assignment}
        \State Assign rewards: $r_{\text{MI}}$ to descriptions, $r_{\text{task}}$ to complete trajectories
        \State Update $\theta$ via policy gradient with decoupled advantages
    \EndFor
\EndFor
\end{algorithmic}
\end{algorithm}

\begin{table*}[t]
\caption{Main performance comparison. Config format: for GRPO, $n$ denotes trajectory count; for \textsc{MIRL}, $N{\to}K{\times}(M+1)$ denotes pre-sample $N$ descriptions, select top-$K$, fork $M$ trajectories each. Best results are \textbf{bolded}, second best are \underline{underlined}.}
\label{tab:main}
\vskip 0.15in
\begin{center}
\begin{small}
\begin{sc}
\begin{tabular}{llccccccc}
\toprule
& & \multicolumn{3}{c}{Math Reasoning} & CoT & \multicolumn{2}{c}{VL} & \\
\cmidrule(lr){3-5} \cmidrule(lr){7-8}
Method & Config & MathVista & MathVerse & We-Math & M3CoT & MMStar & RealWorldQA & Avg. \\
\midrule
Base Model & -- & 69.05 & 35.09 & 47.02 & 63.48 & 54.85 & \underline{59.40} & 54.81 \\
\midrule
DAPO-8 & $n$=8 & 74.95 & 66.27 & 66.05 & 71.10 & 63.73 & 55.49 & 66.27 \\
DAPO-12 & $n$=12 & \textbf{78.63} & \underline{69.52} & 70.68 & 73.74 & \textbf{67.46} & 55.02 & 69.17 \\
DAPO-16 & $n$=16 & 77.89 & 68.12 & \underline{71.20} & \textbf{75.14} & \textbf{67.46} & 58.78 & \underline{69.76} \\
\midrule
MIRL-8 & 8$\to$4$\times$2 & 77.47 & 68.62 & 70.97 & 72.90 & 65.76 & 54.55 & 68.38 \\
MIRL-12 & 10$\to$6$\times$2 & \underline{78.42} & \textbf{70.08} & \textbf{71.54} & \underline{74.35} & \underline{66.58} & \textbf{60.34} & \textbf{70.22} \\
\bottomrule
\end{tabular}
\end{sc}
\end{small}
\end{center}
\vskip -0.1in
\end{table*}

\section{Experiments}
\label{sec:experiments}

In this section, we conducted experiments to address the following research questions:

\begin{itemize}[leftmargin=2em, label=$\bullet$]
    \item \textbf{RQ1}: Can \textsc{MIRL} achieve better performance to uniform sampling baselines while using fewer complete trajectory samples?
    \item \textbf{RQ2}: Is the performance gain truly attributable to MI-guided selection, rather than merely the branching strategy itself?
    \item \textbf{RQ3}: How does the decoupled reward mechanism resolve the credit assignment challenge?
\end{itemize}

\subsection{Experimental Setup}
\label{subsec:setup}

We begin by briefly outlining the evaluation metrics, datasets, and baselines. For more detailed descriptions of the experimental settings, please refer to Appendix~\ref{app:details}.

\paragraph{Models and Baselines.}
Our experiments are conducted on \textbf{Qwen2.5-VL-7B-Instruct}~\citep{DBLP:journals/corr/abs-2409-12191}.
We train on the ViRL39K dataset, which covers diverse vision-language reasoning tasks, formatting responses as \texttt{<description>... <think>... \textbackslash boxed\{ans\}}.
We compare \textsc{MIRL} with the DAPO baseline~\cite{DBLP:journals/corr/abs-2503-14476} 
under different trajectory budgets (\textbf{DAPO-8/12/16}).
For our method, we evaluate \textbf{MIRL-8} (pre-sample 8, select top-4, fork 1) and \textbf{MIRL-12} (pre-sample 10, select top-6, fork 1).

\paragraph{Evaluation Benchmarks and Metrics.}
We evaluate on six benchmarks: \textbf{MathVista}~\citep{DBLP:conf/iclr/LuBX0LH0CG024}, \textbf{MathVerse}~\citep{DBLP:conf/eccv/ZhangJZLGQZLCQGL24}, and \textbf{We-Math} for mathematical reasoning, \textbf{M3CoT} for multi-modal chain-of-thought, \textbf{MMStar}~\citep{DBLP:conf/nips/ChenLDZZCDWQLZ24} and \textbf{RealWorldQA}~\citep{xai2024realworldqa} for general vision-language understanding.
We report \textbf{greedy accuracy} (temperature$=$0); extended results on additional benchmarks are in Appendix~\ref{app:benchmarks}.

\paragraph{Implementation Details}
We implement the DAPO algorithm using the \texttt{verl}~\citep{verl} framework. The model is trained for 2 epochs with a constant learning rate of $1 \times 10^{-6}$. We use a global batch size of 384 and a mini-batch size of 128. During training, we disable the standard KL penalty and enable online filtering with a threshold of $\tau = 0.99$ to select high-quality trajectories for policy updates. The actor clip ratio is set between 0.2 and 0.28.

\begin{figure*}[ht]
  \vskip 0.2in
  \begin{center}
    \centerline{\includegraphics[width=\textwidth]{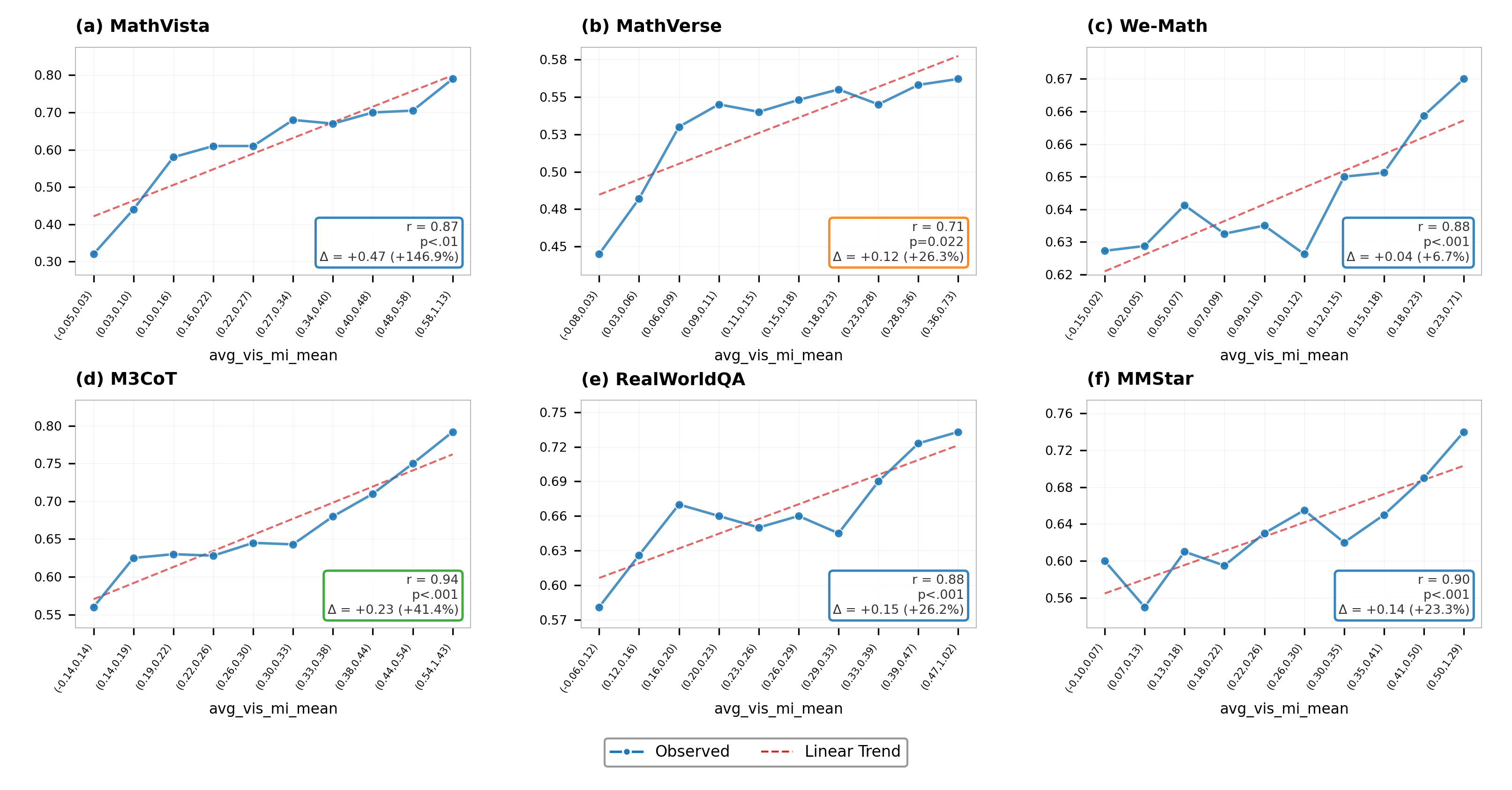}}
    \caption{Vision success rate vs. MI score across six benchmarks. Higher MI correlates with higher task accuracy (Pearson $r=0.71$--$0.94$, all $p<0.05$). \textbf{Samples answerable without visual input are excluded.}}
    \label{fig:mi_correlation}
  \end{center}
  \vskip -0.2in
\end{figure*}

\subsection{Main Results (RQ1)}
\label{subsec:rq1}

To evaluate the performance of different sampling strategies for vision-language RL, we compare \textsc{MIRL} against the DAPO baseline~\cite{DBLP:journals/corr/abs-2503-14476} under various trajectory budgets. DAPO baselines employ standard chain sampling with $n$ independent trajectories. \textsc{MIRL} adopts a fork-branch strategy: pre-sampling $N$ descriptions, selecting top-$K$ by MI scores, and forking each into another $M$ branches, yielding $K{\times}(M+1)$ complete trajectories in total. Based on Table~\ref{tab:main}, we can draw the following observations:

\textbf{Obs 1: MIRL outperforms uniform sampling with fewer complete trajectories.}
\textsc{MIRL}-8, which generates only 8 complete trajectories (4 selected $\times$ 2 branches), achieves 68.38\% average accuracy, comparable to DAPO-12's 69.17\% while using \textbf{33\% fewer complete trajectories} (8 vs. 12). More importantly, \textsc{MIRL}-8 significantly outperforms the sampling-cost-equivalent DAPO-8 (66.27\%) by 2.11 absolute points, demonstrating clear efficiency gains from MI-guided selection. 
Furthermore, \textsc{MIRL}-12 (70.22\%) surpasses DAPO-16 (69.76\%) with 25\% fewer trajectories (12 vs. 16), demonstrating that MI-guided selection effectively concentrates sampling budget on high-potential trajectories.

\textbf{Obs 2: Consistent improvements across diverse task domains.}

\textsc{MIRL}-12 demonstrates robust performance across various reasoning domains. On mathematical reasoning tasks (MathVista, MathVerse, We-Math), it achieves an average score of 73.35\%, surpassing DAPO-16's 72.40\% by 0.95 points. For chain-of-thought reasoning (M3CoT), \textsc{MIRL}-12 reaches 74.35\%, comparable to DAPO-16's 75.14\%. On multimodal understanding benchmarks (MMStar and RealWorldQA), it attains an average of 63.46\%, surpassing DAPO-16's 63.12\% by 0.34 points. Notably, \textsc{MIRL}-12 achieves the best performance on RealWorldQA (60.34\%), addressing superior real-world visual understanding with fewer trajectories. This consistent performance across domains suggests that the decoupling mechanism provides more balanced supervision signals: while the MI reward optimizes visual-linguistic alignment and perception quality, the task reward focuses on reasoning accuracy, enabling the model to develop robust capabilities across diverse task requirements without overfitting to specific benchmark characteristics.
Notably, these improvements are achieved with approximately 14\% 
fewer computational resources compared to DAPO-16 
(see Appendix~\ref{app:cost_analysis} for detailed cost analysis).




\begin{figure}[t]
\begin{center}
\centerline{\includegraphics[width=\columnwidth]{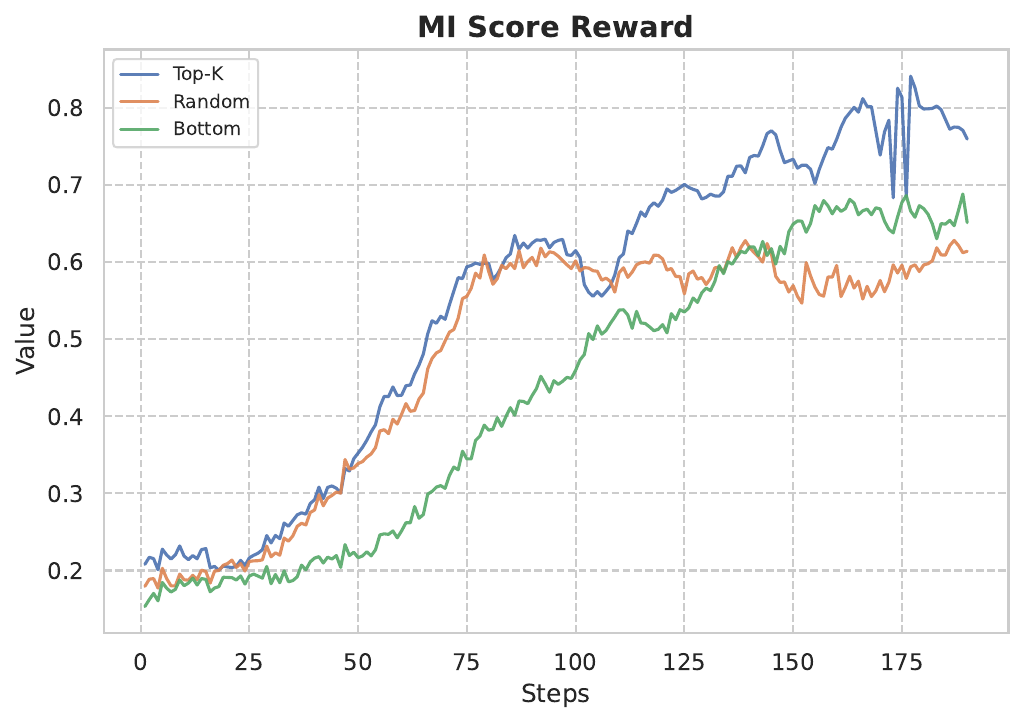}}
\caption{Training dynamics of the MI Score in the ablation on selection strategy.}
\label{fig:strategy_mi_score}
\end{center}
\vskip -0.2in
\end{figure}

\begin{table}[t]
\caption{Ablation on selection strategy. All methods use the same branching configuration (10 pre-samples $\to$ 6 selected $\times$ 2 forks), and this setup is uniformly applied to all subsequent ablation experiments. Random selects uniformly; Bottom selects lowest MI descriptions.}
\label{tab:selection}
\vskip 0.15in
\begin{center}
\begin{small}
\begin{sc}
\begin{tabular}{lccc}
\toprule
Dataset & Random & Bottom & Top-K (Ours) \\
\midrule
Math & 72.58 & 70.75 & \textbf{73.35} \\
CoT & 70.92 & 71.14 & \textbf{74.35} \\
VL & 61.88 & 62.38 & \textbf{63.46} \\
\midrule
Avg. & 68.74 & 68.02 & \textbf{70.22} \\
\bottomrule
\end{tabular}
\end{sc}
\end{small}
\end{center}
\vskip -0.1in
\end{table}

\subsection{MI Attribution Analysis (RQ2)}
\label{subsec:rq3}

A critical question is whether \textsc{MIRL}'s performance gains genuinely stem from MI-guided selection or merely from the branching strategy itself. We design controlled ablations to isolate the causal effect of MI-based selection.

\textbf{Obs 3: MI reliably predicts trajectory success.}
Figure~\ref{fig:mi_correlation} shows that MI exhibits consistent positive correlation with task accuracy across all six benchmarks spanning mathematical reasoning and general visual understanding. On MathVista, accuracy increases from approximately 40\% in the lowest MI quartile to over 70\% in the highest---a relative improvement exceeding 75\%. This validates MI's universal value as a computationally cheap proxy for visual perception quality. 
High MI indicates that the generated description depends on visual information rather than textual priors, which is a prerequisite for correct downstream reasoning.

\textbf{Obs 4: MI selection is the key, not branching alone.}
To isolate the effect of MI-based selection, we compare against Random (same branching but random selection) and Bottom (selecting 6 lowest MI). 
As shown in Table~\ref{tab:selection}, the Top-K strategy consistently outperforms both baselines across all dataset categories.
Random achieves only 72.58\% on Math Benchmarks compared to \textsc{MIRL}-12's 73.35\%, while Bottom performs even worse at 70.75\%. 
Similar patterns emerge on CoT and VL tasks, where Top-K surpasses Random by 3.43\% and 1.58\% respectively, with Bottom again trailing behind both alternatives.
This confirms that performance gains arise specifically from MI-based selection, and that low-MI descriptions introduce noise rather than beneficial diversity.
 
The training dynamics in Figure~\ref{fig:strategy_mi_score} provide additional evidence. The Top-K curve rises steadily and converges to approximately 0.8 after 150 steps. The Random curve, however, oscillates throughout training and plateaus near 0.6. Bottom performs worst, remaining below 0.4 with high variance for most of the training process. These results suggest that high-MI descriptions yield stable learning signals, while low-MI samples introduce noise that impedes effective policy optimization.

\subsection{Decoupled Reward Analysis (RQ3)}
\label{subsec:rq4}

Beyond sampling efficiency, \textsc{MIRL} employs MI-based rewards for description tokens. We investigate whether this decoupled training provides benefits beyond the sampling allocation mechanism.

\begin{figure}[t]
\begin{center}
\centerline{\includegraphics[width=\columnwidth]{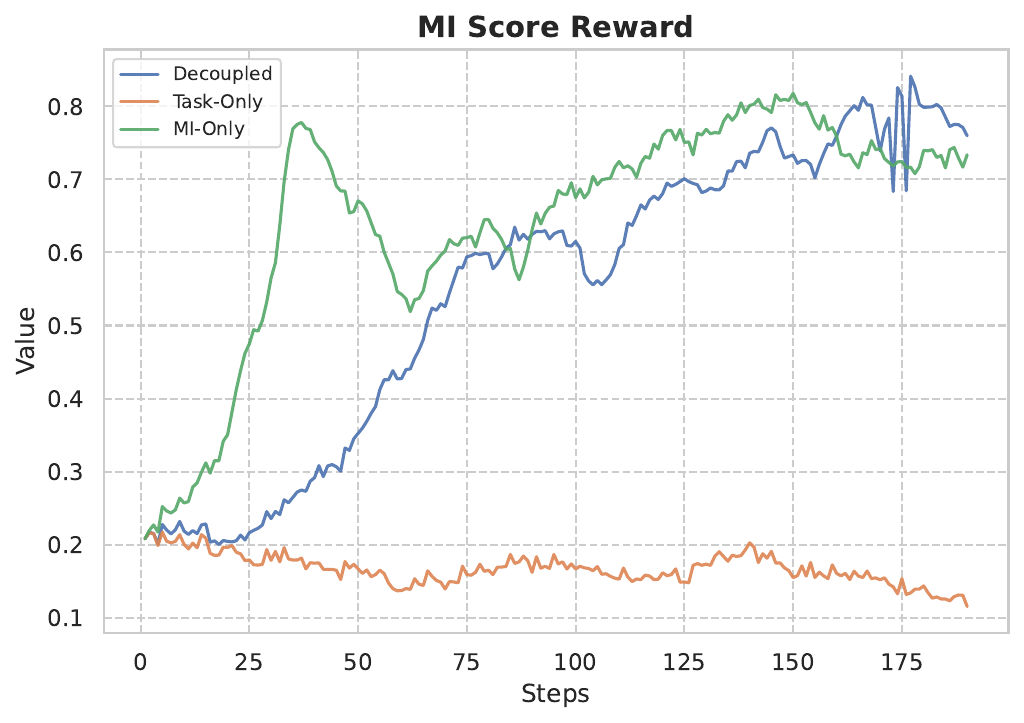}}
\caption{Training dynamics of the MI Score in the ablation on reward configuration.}
\label{fig:reward_mi_score}
\end{center}
\vskip -0.2in
\end{figure}

\begin{table}[t]
\caption{Ablation on reward configuration. ``Decoupled'' (ours) applies MI reward only to description tokens and task reward only to reasoning tokens.}
\label{tab:reward_ablation}
\vskip 0.15in
\begin{center}
\begin{small}
\begin{sc}
\begin{tabular}{lccc}
\toprule
Dataset & Task-only & MI-only & Decoupled \\
MI Wt. & ($\lambda$=0.0) & ($\lambda$=1.0) & (Ours: $\lambda$=0.1) \\
\midrule
Math & 69.22 & 71.84 & \textbf{73.35} \\
CoT & 72.11 & 72.42 & \textbf{74.35} \\
VL & \textbf{65.23} & 62.45 & 63.46 \\
\midrule
Avg. & 68.37 & 68.81 & \textbf{70.22} \\
\bottomrule
\end{tabular}
\end{sc}
\end{small}
\end{center}
\vskip -0.1in
\end{table}

\textbf{Obs 5: Explicit MI supervision promotes visual information utilization.}
As shown in Table~\ref{tab:reward_ablation}, relying solely on sparse task supervision (Task-only, $\lambda=0.0$) yields an average accuracy of 68.37\%. This corresponds to the training dynamics in Figure~\ref{fig:reward_mi_score}, where the Task-only baseline (orange curve) fails to improve the MI score, remaining at a low level throughout training. This indicates that without explicit MI supervision, the model neglects visual information. In contrast, the Decoupled method (blue curve) shows a steady increase in MI score alongside the performance boost to 70.22\%, verifying that MI reward effectively encourages the model to leverage visual contexts to reduce uncertainty, serving as a critical complement to task rewards.

\textbf{Obs 6: Decoupled strategy balances visual grounding and reasoning.}
While MI reward is beneficial, exclusively or heavily prioritizing it (MI-only, $\lambda=1.0$) results in suboptimal performance (68.81\%) compared to our Decoupled configuration (70.22\%). As visualized in Figure~\ref{fig:reward_mi_score}, the MI-only model (green curve) exhibits a rapid and aggressive rise in MI score early in training. However, this aggressive optimization does not translate to the best task accuracy, suggesting the model may over-prioritize visual dependency over reasoning logic. Our Decoupled approach ($\lambda=0.1$, blue curve) demonstrates a more stable and sustainable growth in MI score, confirming that a moderated reward applied specifically to description tokens achieves the best balance between visual grounding and reasoning.

\section{Related Work}
\label{sec:related}

\textbf{Reinforcement Learning with Verifiable Rewards for VLMs.} 
This paradigm optimizes vision-language models using verifiable outcome signals without external supervision. 
GRPO~\citep{DBLP:journals/corr/abs-2402-03300} introduced group-relative policy optimization, eliminating the need for critic models. This has been extended to vision-language domains through VLM-R1~\citep{DBLP:journals/corr/abs-2504-07615} using rule-based rewards on tasks like referring expression comprehension, and Vision-R1~\citep{DBLP:journals/corr/abs-2503-06749} which constructs multimodal chain-of-thought datasets. V-STaR~\citep{DBLP:journals/corr/abs-2402-06457} enhances self-training by leveraging both correct and incorrect solutions through DPO. For finer-grained supervision, process reward methods like Math-Shepherd~\citep{DBLP:conf/acl/WangLSXDLCWS24} and OmegaPRM~\citep{DBLP:journals/corr/abs-2406-06592} evaluate step-wise correctness, with OmegaPRM achieving $75\times$
 efficiency improvement via MCTS-based error identification.

\textbf{Mutual Information in Vision-Language Understanding.}
Mutual information has been widely leveraged for multimodal alignment and quality assessment. CLIP~\citep{DBLP:conf/icml/RadfordKHRGASAM21} maximizes MI through contrastive learning, while InfoNCE~\citep{DBLP:journals/corr/abs-1807-03748} provides tractable bounds adopted in pretraining~\citep{DBLP:conf/icml/ChenK0H20, DBLP:conf/cvpr/He0WXG20}. For hallucination mitigation, M3ID~\citep{DBLP:conf/cvpr/FaveroZTCPASS24} introduces multimodal mutual information decoding to address the ``conditioning dilution'' phenomenon where VLMs over-rely on language priors, penalizing tokens with low visual dependency. VCD~\citep{DBLP:conf/cvpr/LengZCLLMB24} contrasts distributions with/without visual inputs, while OPERA~\citep{DBLP:conf/cvpr/HuangDZ0H0L0Y24} penalizes over-trust in summary tokens during beam search. CMI-VLD~\citep{DBLP:journals/corr/abs-2505-19678} reformulates hallucination as conditional MI maximization with bi-level optimization. These methods operate at inference time, requiring per-token recomputation.

\textbf{Efficient Sampling in Reinforcement Learning.}
Improving sample efficiency remains fundamental to scalable RL. 
Best-of-N sampling~\citep{DBLP:journals/corr/abs-2110-14168} generates multiple candidates and selects via majority voting or reward-based ranking. For reasoning tasks, self-consistency~\citep{DBLP:conf/iclr/0002WSLCNCZ23} aggregates multiple reasoning paths through majority vote, and STaR~\citep{DBLP:conf/nips/ZelikmanWMG22} bootstraps training by filtering self-generated correct solutions. ReST~\citep{DBLP:journals/corr/abs-2308-08998} extends this through iterative refinement with growing datasets. Curriculum learning~\citep{DBLP:conf/icml/BengioLCW09} schedules training examples by predefined difficulty metrics but lacks adaptive mechanisms for dynamic reallocation based on sample quality during training.

\section{Limitations \& Future Work}
\label{sec:limitations}

While \textsc{MIRL} demonstrates significant improvements in sampling efficiency, we acknowledge important limitations. 

\paragraph{Domain Coverage.}
Our experiments focus primarily on mathematical and scientific reasoning with visual diagrams, and the effectiveness of MI-based selection in other vision-language scenarios such as open-ended visual captioning, video understanding, or embodied AI remains to be validated. Additionally, our current implementation focuses on static images, and extending to temporal visual modalities would require adapting MI computation to capture temporal dependencies. Future research could focus on validating \textsc{MIRL}
 across broader multimodal scenarios and exploring its application to dynamic visual content.

\paragraph{Model Generalization.}
Our experiments are conducted primarily on Qwen2.5-VL-7B-Instruct. To verify that the MI-accuracy correlation is not model-specific, we conducted additional validation experiments on Vision-SR1 and GThinker, observing consistent positive correlations across different architectures (see Appendix~\ref{app:cross_model} for details). However, the full \textsc{MIRL} training pipeline has only been evaluated on a single model architecture. Future work should validate \textsc{MIRL}
's effectiveness across diverse VLM architectures (e.g., LLaVA, InternVL) and model scales (from 2B to 70B+ parameters) to establish broader generalizability.

\section{Conclusion}
\label{sec:conclusion}

In this work, we introduced \textsc{MIRL}, a decoupled training framework that addresses two critical bottlenecks in vision-language reinforcement learning: poor sampling efficiency and reward blindness to visual grounding quality.
Specifically, \textsc{MIRL} leverages mutual information between generated descriptions and visual inputs as a cheap pre-screening signal, enabling intelligent budget reallocation toward high-potential descriptions via forking.
Then, the decoupled training provides immediate MI-based rewards for visual perception while task rewards supervise reasoning.
Extensive experiments on multiple benchmarks spanning mathematical reasoning to general vision-language understanding demonstrate that \textsc{MIRL} achieves better performance compared to uniform sampling with 16 complete trajectories while using only 10 pre-samples with top-6 selection, reducing computational cost by approximately 25\%.






\section*{Reproducibility}

To ensure reproducibility of our findings, we provide:

Detailed reproduction instructions, including environment setup, dependency installation, 
and step-by-step commands, are provided in the README of our code repository.



\bibliography{example_paper}
\bibliographystyle{icml2026}

\newpage
\appendix
\onecolumn


\section{Experimental Details}
\label{app:details}

\subsection{Training Configuration}
\label{app:training_config}

We conduct all experiments on \textbf{Qwen2.5-VL-7B-Instruct}~\citep{DBLP:journals/corr/abs-2409-12191} using 8 NVIDIA A100 (80GB) GPUs. The training hyperparameters are summarized in Table~\ref{tab:hyperparams}.

\begin{table}[h]
\caption{Training hyperparameters for MIRL experiments.}
\label{tab:hyperparams}
\vskip 0.15in
\begin{center}
\begin{small}
\begin{sc}
\begin{tabular}{lc}
\toprule
Hyperparameter & Value \\
\midrule
Learning Rate & $1 \times 10^{-6}$ \\
Weight Decay & $1 \times 10^{-2}$ \\
Optimizer & AdamW \\
LR Warmup Ratio & 0.0 \\
Max Grad Norm & 1.0 \\
KL Coefficient & $1 \times 10^{-2}$ \\
Rollout Temperature & 1.0 \\
Top-p Sampling & 0.99 \\
Max Prompt Length & 4096 \\
Max Response Length & 2048 \\
Global Batch Size & 128 \\
Rollout Batch Size & 384 \\
Total Epochs & 2--3 \\
\bottomrule
\end{tabular}
\end{sc}
\end{small}
\end{center}
\vskip -0.1in
\end{table}

\paragraph{Dataset Composition.}
 ViRL39K covers five primary categories: (1) mathematical reasoning with visual diagrams (geometry, charts, graphs), (2) scientific figure comprehension, (3) document and table understanding, (4) spatial reasoning and object relationships, and (5) general visual question answering requiring multi-step inference. Each sample includes an image, a textual query, and a verifiable ground-truth answer suitable for automatic reward computation.

\paragraph{Quality Filtering.}
 The dataset undergoes rigorous filtering to ensure: (a) questions genuinely require visual information (text-only answerable samples are removed), (b) answers are unambiguous and automatically verifiable, and (c) reasoning complexity is sufficient to benefit from RL optimization rather than simple pattern matching.

\paragraph{Format Standardization.} All samples are formatted to elicit structured responses following the \texttt{<description>...</description><think>...</think>\textbackslash boxed\{answer\}}
 template, enabling clean separation between visual perception and reasoning stages for our decoupled training approach.

\subsection{MI Score Computation Details}

The mutual information between generated tokens and visual input is computed as follows. For each token $y_t$
 in the description segment, we compute:
\begin{equation}
MI(y_t; I | q, y_{<t}) = H(y_t | q, y_{<t}) - H(y_t | I, q, y_{<t}),
\end{equation}
where the conditional entropy $H(y_t | q, y_{<t})$ is estimated by performing a forward pass without the image input, and $H(y_t | I, q, y_{<t})$
 is estimated with the image input.

In practice, we compute:
\begin{align}
H(y_t | q, y_{<t}) &= -\log p_\theta(y_t | q, y_{<t}) \\
H(y_t | I, q, y_{<t}) &= -\log p_\theta
(y_t | I, q, y_{<t})
\end{align}

The MI score is then clipped to $[0, 0.5]$
 and normalized for reward computation:
\begin{equation}
r_{MI} = \frac{\max(0, \min
(0.5, MI))}{0.5}
\end{equation}

\subsection{Description Segment Identification}
\label{app:desc_identification}

Our method requires identifying the description segment $D$
 from model responses for MI computation and decoupled training. We employ different strategies depending on the model's output format:

\paragraph{For models with explicit format tags.} When models are trained to produce structured outputs with explicit delimiters (e.g., \texttt{<description>...</description>}
 tags), we directly extract content within these tags. This provides clean and unambiguous segment boundaries, ensuring robust identification across different response lengths and styles.

\paragraph{For models without explicit format tags.} We segment responses using double newlines (\texttt{"\textbackslash n\textbackslash n"}
) as natural paragraph boundaries and compute MI for each segment independently. The segment with the highest aggregated MI score is selected as the description, based on the intuition that visual perception content exhibits stronger dependence on image input than pure reasoning text.

\paragraph{Validation of automatic identification.} To verify that MI-based segment selection genuinely captures visual recognition content, we employed GPT-4 as an evaluator on 140 rollout samples. The evaluation prompt asked whether the extracted segment performs ``initial diagram information recognition''---defined as the first instance where information is explicitly read or described from the visual input. GPT-4 evaluation achieved \textbf{74\% agreement} with our MI-based selection, confirming that high-MI segments predominantly correspond to genuine visual perception rather than reasoning or answer formulation stages.

\subsection{Response Format}
\label{app:response_format}

We train models to generate responses in the following structured format:
\begin{verbatim}
<description> visual description </description>
<think> reasoning process </think>
\boxed{final answer}
\end{verbatim}

The \texttt{<description>} block contains the visual perception output that describes task-relevant visual information from the image. The \texttt{<think>} block contains the step-by-step reasoning process. The final answer is enclosed in \texttt{\textbackslash boxed\{\}}.

\section{Benchmark Descriptions}
\label{app:benchmarks}

We evaluate \textsc{MIRL} on seven diverse benchmarks spanning mathematical reasoning, multimodal chain-of-thought, general vision-language understanding, logical reasoning, and hallucination detection. Below we provide detailed descriptions of each benchmark.

\subsection{MathVista}
\label{app:mathvista}

\textbf{MathVista}~\citep{DBLP:conf/iclr/LuBX0LH0CG024} is a comprehensive benchmark designed to evaluate mathematical reasoning capabilities of foundation models in visual contexts. It consists of 6,141 examples derived from 28 existing multimodal datasets involving mathematics and 3 newly created datasets: IQTest, FunctionQA, and PaperQA. These newly created datasets address missing visual domains and are tailored to evaluate logical reasoning on puzzle test figures, algebraic reasoning over functional plots, and scientific reasoning with academic paper figures, respectively.

MathVista encompasses seven mathematical reasoning types: algebraic reasoning, arithmetic reasoning, geometry reasoning, logical reasoning, numeric commonsense, scientific reasoning, and statistical reasoning. The benchmark includes five primary task types: figure question answering (FQA), geometry problem solving (GPS), math word problems (MWP), textbook question answering (TQA), and visual question answering (VQA).

The dataset is split into two subsets: \textbf{testmini} (1,000 examples) for model development and validation, and \textbf{test} (5,141 examples) for standard evaluation. We use the testmini subset for our experiments.

\subsection{MathVerse}
\label{app:mathverse}

\textbf{MathVerse}~\citep{DBLP:conf/eccv/ZhangJZLGQZLCQGL24} is an all-around visual math benchmark specifically designed for equitable and in-depth evaluation of multimodal large language models (MLLMs). The benchmark addresses a critical limitation of existing benchmarks: they often incorporate excessive visual content within textual questions, which may assist MLLMs in deducing answers without truly interpreting the input diagrams.

MathVerse comprises 2,612 high-quality, multi-subject math problems with diagrams collected from publicly available sources. Each problem is transformed by human annotators into six distinct versions, each offering varying degrees of information content in multi-modality, contributing to 15K test samples in total. The benchmark spans three primary mathematical areas: plane geometry, solid geometry, and functions. Each problem is classified into 12 detailed categories emphasizing different fine-grained problem-solving capabilities.

A key innovation of MathVerse is its multi-version transformation strategy that creates six versions of each problem: Text Dominant, Text Lite, Vision Intensive, Vision Dominant, Vision Only, and Text Only. This design allows comprehensive assessment of whether and how much MLLMs can truly understand visual diagrams for mathematical reasoning.

\subsection{We-Math}
\label{app:wemath}

\textbf{We-Math}~\citep{DBLP:conf/acl/QiaoTDWSSWGLZWZ25} is the first benchmark specifically designed to explore problem-solving principles beyond end-to-end performance in visual mathematical reasoning. Inspired by human-like mathematical reasoning processes, We-Math meticulously collects and categorizes 6.5K visual math problems spanning 67 hierarchical knowledge concepts and 5 layers of knowledge granularity.

A distinctive feature of We-Math is its knowledge-based decomposition approach: composite problems are decomposed into sub-problems according to required knowledge concepts. The benchmark introduces a novel four-dimensional evaluation metric:
\begin{itemize}[leftmargin=2em]
    \item \textbf{Insufficient Knowledge (IK)}: The model lacks necessary knowledge concepts.
    \item \textbf{Inadequate Generalization (IG)}: The model possesses knowledge but fails to generalize.
    \item \textbf{Complete Mastery (CM)}: The model correctly solves both composite and sub-problems.
    \item \textbf{Rote Memorization (RM)}: The model solves composite problems but fails on sub-problems.
\end{itemize}

This hierarchical assessment reveals inherent issues in LMMs' reasoning processes and provides insights into knowledge acquisition and generalization capabilities.

\subsection{M3CoT}
\label{app:m3cot}

\textbf{M3CoT}~\citep{DBLP:conf/acl/Chen0ZC0C24} (Multi-Domain Multi-step Multi-modal Chain-of-Thought) is a novel benchmark that addresses limitations in existing multimodal chain-of-thought benchmarks. Previous MCoT benchmarks face three challenges: (1) absence of visual modal reasoning, (2) single-step visual modal reasoning, and (3) domain missing.

M3CoT advances the field by providing:
\begin{itemize}[leftmargin=2em]
    \item \textbf{Multi-domain coverage}: Problems span diverse domains requiring different types of knowledge.
    \item \textbf{Multi-step reasoning}: Problems require chaining multiple reasoning steps.
    \item \textbf{Multi-modal integration}: Solutions necessitate leveraging both textual and visual modalities.
\end{itemize}

The benchmark reveals that current Vision Large Language Models (VLLMs) still struggle to correctly reason in multi-step, multi-modal scenarios, with a substantial gap between VLLMs and human performance.

\subsection{MMStar}
\label{app:mmstar}

\textbf{MMStar}~\citep{DBLP:conf/nips/ChenLDZZCDWQLZ24} is an elite vision-indispensable multi-modal benchmark comprising 1,500 challenge samples meticulously selected by humans. The benchmark addresses two critical issues identified in existing LVLM evaluation:

\begin{enumerate}[leftmargin=2em]
    \item \textbf{Visual content is unnecessary for many samples}: Answers can be directly inferred from questions and options, or from world knowledge embedded in LLMs.
    \item \textbf{Unintentional data leakage}: LLMs and LVLMs can answer visual-necessary questions without visual content due to memorization during training.
\end{enumerate}

MMStar evaluates 6 core capabilities: Coarse Perception (CP), Fine-grained Perception (FP), Instance Reasoning (IR), Logical Reasoning (LR), Science \& Technology (ST), and Mathematics (MA). These are further divided into 18 detailed axes. Each capability contains a balanced 250 samples, with relatively even distribution across detailed axes.

The benchmark also introduces two novel metrics: \textbf{Multi-modal Gain (MG)} measures actual performance gain from multi-modal training, and \textbf{Multi-modal Leakage (ML)} quantifies data leakage degree.

\subsection{RealWorldQA}
\label{app:realworldqa}

\textbf{RealWorldQA}~\citep{xai2024realworldqa} is a benchmark introduced by xAI to evaluate the real-world spatial understanding capabilities of Multimodal Large Language Models (MLLMs). While current frontier models excel in digital-native tasks or document understanding, they often struggle with basic physical reasoning and spatial awareness in the three-dimensional world, which are essential for developing useful real-world AI assistants.

RealWorldQA specifically evaluates a model's ability to perceive and reason about the physical environment through several dimensions:
\begin{itemize}[leftmargin=2em]
    \item \textbf{Spatial Reasoning}: Identifying relative positions, distances, and orientations of objects in 3D space.
    \item \textbf{Situational Awareness}: Understanding complex real-world scenes, such as identifying traffic directions from a vehicle's perspective.
    \item \textbf{Object Recognition under Variation}: Perceiving objects accurately despite occlusions, diverse lighting, and various camera angles.
    \item \textbf{Physical Commonsense}: Reasoning about the physical constraints and properties of objects in everyday settings.
\end{itemize}

The initial release of the benchmark comprises 763 high-resolution images, primarily consisting of anonymized images taken from vehicles (dashcam views) alongside various household and real-world scenes. Each image is paired with a question and a verifiable answer, providing a challenging and grounded testbed for visual perception.





\section{Algorithm Details}
\label{app:algorithm}

\subsection{Baseline: GRPO and DAPO}
\label{app:baselines}

\textbf{GRPO} (Group Relative Policy Optimization)~\citep{DBLP:journals/corr/abs-2402-03300} is a reinforcement learning algorithm that eliminates the need for critic models by computing advantages relative to other samples in the same group. For each query, $G$ responses are sampled, and advantages are computed as:
\begin{equation}
\hat{A}^{(i)} = \frac{r^{(i)} - \mu_G}{\sigma_G}
\end{equation}
where $\mu_G$ and $\sigma_G$ are the mean and standard deviation of rewards within the group.

\textbf{DAPO}~\citep{DBLP:journals/corr/abs-2503-14476} extends GRPO with several improvements including dynamic clipping ratios and online filtering mechanisms. We use DAPO as our primary baseline for comparison.

\subsection{MIRL Sampling Strategy}
\label{app:mirl_sampling}

The MIRL sampling strategy consists of three phases:

\textbf{Phase 1 (Pre-sampling)}: Generate $N$ response candidates, each containing a description segment $\mathcal{D}^{(i)}$.

\textbf{Phase 2 (MI-based Selection)}: Compute MI scores for all descriptions and select top-$K$:
\begin{equation}
\mathcal{S}_K = \textsc{TopK}\bigl(\{\text{MI}(\mathcal{D}^{(i)}; I \mid q)\}_{i=1}^{N}, K\bigr)
\end{equation}

\textbf{Phase 3 (Forking)}: For each selected description $i \in \mathcal{S}_K$, generate $M$ additional reasoning branches:
\begin{equation}
\tilde{\mathcal{R}}^{(i,j)}, \tilde{a}^{(i,j)} \sim \pi_\theta(\cdot \mid I, q, \mathcal{D}^{(i)}), \quad j = 1, \ldots, M
\end{equation}

This results in $K(M+1)$ complete trajectories for policy updates.

\subsection{Reward Function Details}
\label{app:reward_details}

The reward function combines multiple components:

\textbf{Format Reward}: Checks if the response follows the expected structure:
\begin{equation}
r_{\text{format}} = \mathbb{I}[\text{response matches } \texttt{<description>...<think>...\textbackslash boxed\{...\}}]
\end{equation}

\textbf{Accuracy Reward}: Evaluates answer correctness:
\begin{equation}
r_{\text{acc}} = \mathbb{I}[\texttt{extract\_boxed}(y) = a^*]
\end{equation}

\textbf{MI Reward}: Measures visual grounding quality (clipped and normalized):
\begin{equation}
r_{\text{MI}} = \frac{\max(0, \min(0.5, \text{MI}(\mathcal{D}; I \mid q)))}{0.5}
\end{equation}

\textbf{Overall Reward for Reasoning}: Combines accuracy and format:
\begin{equation}
r_{\text{reasoning}} = (1 - \lambda_f) \cdot r_{\text{acc}} + \lambda_f \cdot r_{\text{format}}
\end{equation}
where $\lambda_f = 0.1$ is the format weight.

\textbf{Overall Reward for Description}: Combines accuracy and MI:
\begin{equation}
r_{\text{description}} = (1 - \lambda_m) \cdot r_{\text{acc}} + \lambda_m \cdot r_{\text{MI}}
\end{equation}
where $\lambda_m$ is the MI weight (default 0.1).

\subsection{Computational Cost Analysis}
\label{app:cost_analysis}

Table~\ref{tab:cost_analysis} provides a detailed breakdown of computational costs for different configurations. We estimate costs in terms of equivalent complete trajectory generations.

\begin{table}[h]
\centering
\caption{Computational cost breakdown. Description segments constitute approximately 15\% of total response length. MI computation requires two forward passes over description tokens only.}
\label{tab:cost_analysis}
\begin{tabular}{lccc}
\toprule
\textbf{Component} & \textbf{DAPO-16} & \textbf{MIRL-8} & \textbf{MIRL-12} \\
 & & (8$\rightarrow$4$\times$2) & (10$\rightarrow$6$\times$2) \\
\midrule
Pre-sampling (desc. only) & 0 & $8 \times 0.15 = 1.2$ & $10 \times 0.15 = 1.5$ \\
MI computation (2 fwd passes) & 0 & $\sim$0.2 & $\sim$0.3 \\
Complete trajectories & 16 & 8 & 12 \\
\midrule
\textbf{Total equiv. trajectories} & \textbf{16.0} & \textbf{9.4} & \textbf{13.8} \\
\textbf{Relative cost} & 100\% & 58.8\% & 86.3\% \\
\midrule
\textbf{Avg. Accuracy (\%)}& 69.76 & 68.38 & \textbf{70.22} \\
\bottomrule
\end{tabular}
\end{table}

\paragraph{Cost estimation details.}
\begin{itemize}[leftmargin=*,nosep]
    \item \textbf{Pre-sampling cost}: Generating only the description segment requires approximately 15\% of the compute needed for a complete trajectory, as descriptions typically comprise 15\% of total response tokens.
    \item \textbf{MI computation cost}: Computing MI requires two forward passes (with and without image) over the description tokens. Since this operates only on the short description segment, the overhead is minimal ($\sim$2\% per description).
    \item \textbf{Forking cost}: Forked trajectories start from the end of the description, so they generate only the reasoning and answer portions ($\sim$85\% of a complete trajectory). However, for simplicity, we count each forked trajectory as one complete trajectory in our analysis, making our cost estimates conservative.
\end{itemize}

\paragraph{Efficiency gains.} MIRL-12 achieves \textbf{higher accuracy} (70.22\% vs. 69.76\%) than DAPO-16 while using approximately \textbf{14\% fewer computational resources}. MIRL-8 achieves comparable performance to DAPO-12 with approximately \textbf{22\% cost reduction}.

\section{Qualitative Examples}
\label{app:qualitative}

\subsection{MI-Guided Selection Examples}
\label{app:selection_examples}

We provide qualitative examples demonstrating how MI scores effectively distinguish high-quality visual descriptions from low-quality ones. 

\textbf{High-MI Description Example} (MI=0.42):
\begin{quote}
\textit{``The image shows a geometric figure with triangle ABC where angle BAC is marked as 35 degrees. Point D lies on side BC, and line segment AD is drawn. The figure indicates that AD is perpendicular to BC, forming a right angle at D.''}
\end{quote}

\textbf{Low-MI Description Example} (MI=0.08):
\begin{quote}
\textit{``This appears to be a math problem involving triangles. There are some angles and lines shown in the figure.''}
\end{quote}

The high-MI description captures specific visual details (angle measurements, perpendicularity) necessary for solving the problem, while the low-MI description relies on generic language patterns without extracting task-relevant visual information.

\subsection{Failure Mode Analysis}
\label{app:failure_modes}

We categorize common failure modes observed during evaluation:

\textbf{Type 1: Visual Perception Failure}
\begin{itemize}[leftmargin=2em]
    \item Missing critical visual elements (e.g., failing to identify angle markers)
    \item Incorrect spatial relationship interpretation
    \item OCR errors in reading numbers or labels
\end{itemize}

\textbf{Type 2: Reasoning Failure}
\begin{itemize}[leftmargin=2em]
    \item Correct description but incorrect theorem application
    \item Arithmetic errors during calculation
    \item Logical inconsistencies in multi-step reasoning
\end{itemize}

\textbf{Type 3: Format Failure}
\begin{itemize}[leftmargin=2em]
    \item Missing required tags (\texttt{<description>}, \texttt{<think>})
    \item Incorrect answer formatting (missing \texttt{\textbackslash boxed\{\}})
\end{itemize}

The decoupled training in MIRL specifically addresses Type 1 failures by providing independent MI-based supervision for the visual perception stage.

\section{Extended Discussion on Limitations}
\label{app:limitations}

\subsection{Cross-Model Validation of MI as Quality Indicator}
\label{app:cross_model}

To verify that MI genuinely measures visual grounding quality rather than exploiting model-specific artifacts, we conducted validation experiments across multiple model architectures beyond our primary training model (Qwen2.5-VL-7B-Instruct).

\paragraph{Vision-SR1 Validation.}
We evaluated on Vision-SR1, which uses explicit \texttt{<visual\_perception>} tags to separate visual description from reasoning. Across all six benchmarks, we observed consistent positive correlations between description MI and task accuracy (Figure~\ref{fig:mi_correlation}), with Pearson coefficients ranging from $r=0.71$ to $r=0.94$
.

\paragraph{GThinker Validation.}
We additionally validated on GThinker, which employs \texttt{<vcues>}
 tags to mark visual reasoning segments. Despite architectural differences in how visual information is structured, we observed similar MI-accuracy correlations, confirming that MI captures a fundamental property of visual grounding quality rather than model-specific behavior.

\paragraph{Filtering Text-Only Answerable Samples.}
During analysis, we discovered that certain benchmark samples can be answered correctly through textual reasoning alone, without genuine visual understanding. For example, on MathVerse, questions providing coordinate information in text (e.g., ``a curve passes through (-1,0), (0,8), (2,0), (4,0)'') enable correct answers without visual inspection. After filtering such text-only answerable samples, the MI-accuracy correlation strengthens consistently across benchmarks, confirming that MI specifically measures visual dependency rather than general response quality.

\paragraph{Implications.}
These cross-model validation results suggest that MI serves as a universal proxy for visual grounding quality, supporting the generalizability of \textsc{MIRL}'s core mechanism. However, we note that the full training pipeline was only evaluated on Qwen2.5-VL-7B-Instruct, and future work should verify that the efficiency gains transfer to other architectures.

\subsection{Domain Generalization}
\label{app:domain_generalization}

While MI demonstrates strong predictive power across the evaluated benchmarks, we acknowledge that its effectiveness may vary in domains where:
\begin{itemize}[leftmargin=2em]
    \item Visual content is highly abstract or symbolic
    \item Questions require minimal visual information
    \item Visual and textual information are redundant
\end{itemize}

Future work should investigate MI's utility in broader visual reasoning scenarios including video understanding, embodied AI, and medical imaging.

\subsection{Computational Considerations}
\label{app:computational}

The current implementation requires two forward passes for MI computation. While this overhead is relatively small compared to complete trajectory generation, it may become significant in extremely large-scale training scenarios. Potential optimizations include:
\begin{itemize}[leftmargin=2em]
    \item Caching intermediate representations for the image-free forward pass
    \item Approximating MI using lightweight proxy models
    \item Adaptive MI computation based on description length
\end{itemize}

\subsection{Hyperparameter sensitivity.} Our experiments use fixed values for key hyperparameters ($N$, $K$, $M$, $\lambda_m$
). While we demonstrate effectiveness with the reported configurations, a comprehensive hyperparameter sensitivity analysis would strengthen the practical applicability of MIRL across different settings and resource constraints.


\end{document}